\documentclass[11pt,a4paper]{article}
\usepackage[hyperref]{emnlp2020}
\usepackage{times}
\usepackage{latexsym}
\usepackage{CJKutf8}

\usepackage{graphicx}
\usepackage{multirow}
\usepackage{amsmath}
\usepackage{float}
\usepackage{subfigure}
\usepackage{algorithm}
\usepackage{algpseudocode}

\newcommand{\tabincell}[2]{\begin{tabular}{@{}#1@{}}#2\end{tabular}} 

\usepackage{microtype}

\aclfinalcopy 
\def\aclpaperid{179} 

\title{Semantically-Aligned Universal Tree-Structured Solver for Math Word Problems}
\author{Jinghui Qin\textsuperscript{\rm 1}, Lihui Lin\textsuperscript{\rm 2}, Xiaodan Liang\textsuperscript{\rm 1,2,}\thanks{\ \ Corresponding Author}\, , Rumin Zhang\textsuperscript{\rm 2}, Liang Lin\textsuperscript{\rm 1,2}\\
\textsuperscript{\rm 1} Sun Yat-sen University \\
\textsuperscript{\rm 2} Dark Matter AI Inc.\\
\texttt{qinjingh@mail2.sysu.edu.cn}, \texttt{linlh23@mail2.sysu.edu.cn}, \\
\texttt{xdliang328@gmail.com}, \texttt{rm\_zhang@foxmail.com}, \\
\texttt{linliang@ieee.org}
}

\date{}

\begin{document}
\maketitle
\begin{abstract}
A practical automatic textual math word problems (MWPs) solver should be able to solve various textual MWPs while most existing works only focused on one-unknown linear MWPs. Herein, we propose a simple but efficient method called Universal Expression Tree (UET) to make the first attempt to represent the equations of various MWPs uniformly. Then a semantically-aligned universal tree-structured solver (SAU-Solver) based on an encoder-decoder framework is proposed to resolve multiple types of MWPs in a unified model, benefiting from our UET representation. Our SAU-Solver generates a universal expression tree explicitly by deciding which symbol to generate according to the generated symbols' semantic meanings like human solving MWPs. Besides, our SAU-Solver also includes a novel subtree-level semantically-aligned regularization to further enforce the semantic constraints and rationality of the generated expression tree by aligning with the contextual information. Finally, to validate the universality of our solver and extend the research boundary of MWPs, we introduce a new challenging \textbf{H}ybrid \textbf{M}ath \textbf{W}ord \textbf{P}roblems dataset (HMWP), consisting of three types of MWPs. Experimental results on several MWPs datasets show that our model can solve universal types of MWPs and outperforms several state-of-the-art models\footnote{The code and the new HMWP dataset are available at \url{https://github.com/QinJinghui/SAU-Solver}.}.
\end{abstract}

\section{Introduction}
Math word problems (MWPs) solving aims to automatically answer a math word problem by understanding the textual description of the problem and reasoning out the underlying answer. A typical MWP is a short story that describes a partial state of the world and poses a question about an unknown quantity or multiple unknown quantities. Thus, a machine should have the ability of natural language understanding and reasoning. To solve an MWP, the relevant quantities need to be identified from the text, and the correct operators and their computation order among these quantities need to be determined.  
\begin{figure}[t] 
	\centerline{\includegraphics[width=0.99\linewidth]{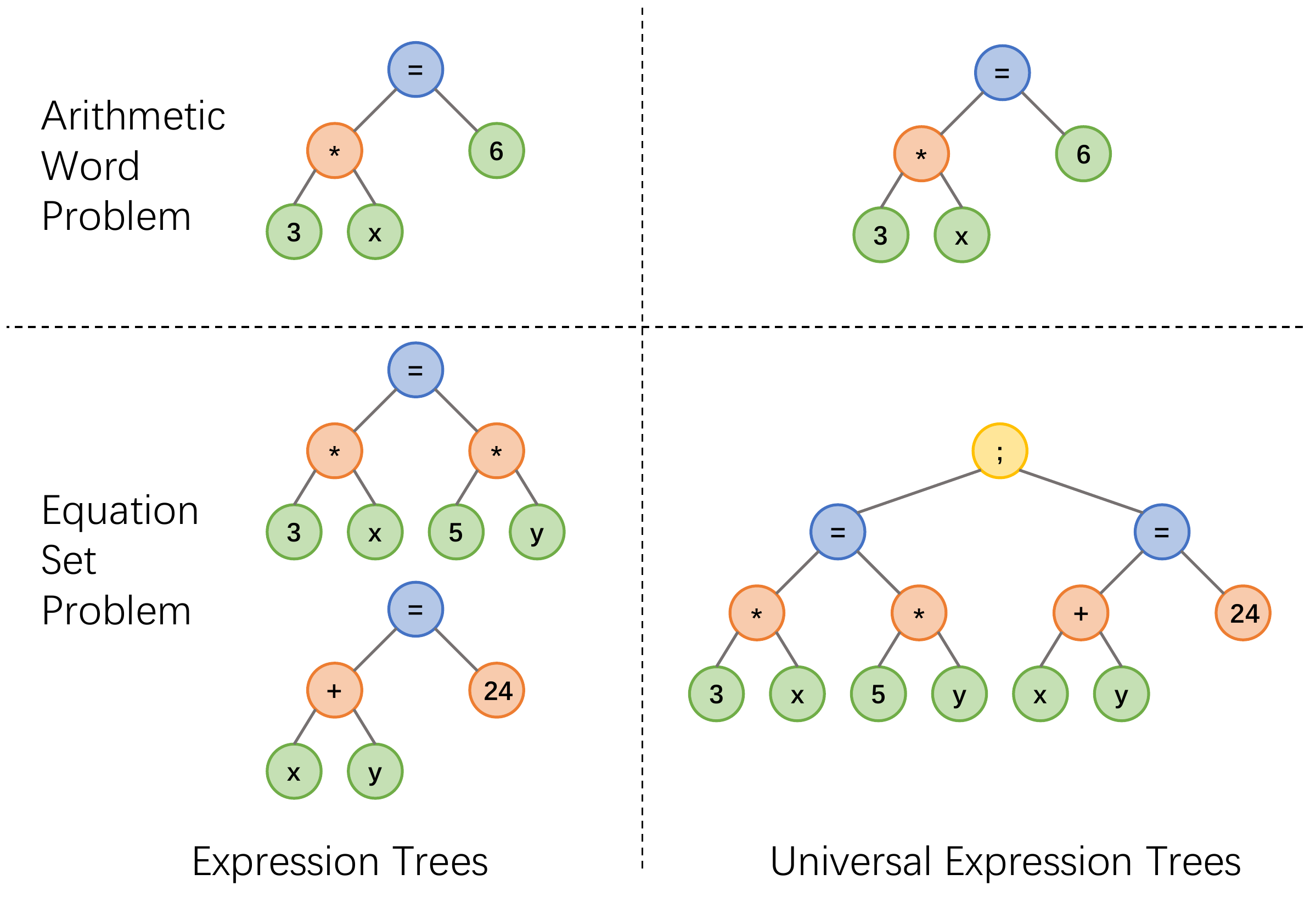}}
	\caption{Universal Expression Trees (UET). In our UET representation, multiple expression trees underlying a MWP will be integrated as an universal expression tree (UET) via symbol extension. UET can enable a solver to handle multiple types of MWPs in an unified manner like a single expression tree of an equation.}
	\label{fig:uet}
\end{figure}

Many traditional methods~\cite{yuhui2010frame-based,kushman2014learning,shi-etal-2015-automatically} have been proposed to address this problem, but they relied on tedious hand-crafted features and template annotation, which required extensive human efforts and knowledge.
Recently, deep learning has opened a new direction towards automatic MWPs solving~\cite{dns,cass,mathdqn,trnn,seq2tree,stackdecoder}. Most of deep learning-based methods try to train an end-to-end neural network to automatically learn the mapping function between problems and their corresponding equations. However, there are some limitations hindering them from being applied in real-world applications.  First, although seq2seq model~\cite{dns} can be applied to solve various MWPs, it suffers from fake numbers generation and mispositioned numbers generation due to all data share the same target vocabulary without problem-specific constraints. Second, some advanced methods~\cite{mathdqn,trnn,seq2tree} only target at arithmetic word problems without any unknown or with one unknown that do not need to model the unknowns underlying in MWPs, which prevent them from generalizing to various MWPs, such as equation set problems. Thus, their methods can only handle arithmetic problems with no more than one unknown. Besides, they also lack an efficient equation representation mechanism to handle those MWPs with multiple unknowns and multiple equations, such as equation set problems. Finally, though some methods~\cite{dns,cass,stackdecoder} can handle multiple types of MWPs, they neither generate next symbol by taking full advantage of the generated symbols like a human nor consider the semantic transformation between equations in a problem, resulting in poor performance on the multiple-unknown MWPs, such as the MWPs involving equation set.

To address the above issues, we propose a simple yet efficient method called Universal Expression Tree (UET) to make the first attempt to represent the equations of various MWPs uniformly like the expression tree of one-unknown linear word problems with considering unknowns. Specifically, as shown in Fig.~\ref{fig:uet}, UET integrates all expression trees underlying in an MWP into an ensemble expression tree via math operator symbol extension so that the grounded equations of various MWPs can be handled in a unified manner as handling one-unknown linear MWPs. Thus, it can significantly reduce the difficulty of modeling equations of various MWPs. 

Then, we propose a semantically-aligned universal tree-structured solver (SAU-Solver), which is based on our UET representation and an Encoder-Decoder framework, to solve multiple types of MWPs in a unified manner with a single model. In our SAU-Solver, the encoder is designed to understand the semantics of MWPs and extract number semantic representation while the tree-structured decoder is designed to generate the next symbol based on the problem-specific target vocabulary in a semantically-aligned manner by taking full advantage of the semantic meanings of the generated expression tree like a human uses problem's contextual information and all tokens written to reason next token for solving MWPs. The problem-specific target vocabulary can help our solver to mitigate the problem of fake numbers generation as much as possible.

Besides, to further enforce the semantic constraints and rationality of the generated expression tree, we also propose a subtree-level semantically-aligned regularization to further improve subtree-level semantic representation by aligning with the contextual information of a problem, which can improve answer accuracy effectively. 

Finally, to validate the universality of our solver and push the research boundary of MWPs to math real-word applications better, we introduce a new challenging \textbf{H}ybrid \textbf{M}ath \textbf{W}ord \textbf{P}roblems dataset (HMWP), consisting of one-unknown linear word problems, one-unknown non-linear word problems, and equation set problems with two unknowns. Experimental results on HWMP, ALG514, Math23K, and Dolphin18K-Manual show the universality and superiority of our approach compared with several state-of-the-art methods. 

\section{Related Works}
Numerous methods have been proposed to attack the MWPs task, ranging from rule-based methods~\cite{ bakman2007robust, yuhui2010frame-based}, statistical machine learning methods~\cite{kushman2014learning, zhou-etal-2015-learn, mitra-baral-2016-learning, huang-etal-2016-well, roy2018mapping},semantic parsing methods~\cite{shi-etal-2015-automatically,koncelkedziorski2015parsing,huang-etal-2017-learning}, and deep learning methods~\cite{ling-etal-2017-program, dns, mathdqn, cass, seq2et, seq2tree, trnn}. Due to space limitations, we only review some recent advances on deep leaning-based methods. ~\cite{dns} made the first attempt to generate expression templates using Seq2Seq model. Seq2seq method has achieved promising results, but it suffers from generating spurious numbers, predicting numbers at wrong positions, or equation duplication problem~\cite{cass, seq2et}. To address them,~\cite{cass} proposed to add a copy-and-alignment mechanism to the standard Seq2Seq model. ~\cite{seq2et} proposed equation normalization to normalize the duplicated equations by considering the uniqueness of an expression tree. 

Different from seq2seq-based works, ~\cite{seq2tree} proposed a tree-structured decoder to generate an expression tree inspired by the goal-driven problem-solving mechanism. ~\cite{trnn} proposed a two-stage template-based solution based on a recursive neural network for math expression construction. However, they do not model the unknowns underlying in MWPs, resulting in only handling one-unknown linear word problems. Besides, they also lack an efficient mechanism to handle those MWPs with multiple unknowns and multiple equations, such as equation set problems. Therefore, their solution can not solve other types of MWPs that are more challenging due to larger search space, such as equation set problems, non-linear equation problems, etc. ~\cite{stackdecoder} is a general equation generator that generates expression via the stack, but they did not consider the semantic transformation between equations in a problem, resulting in poor performance on the multiple-unknown MWPs, such as equation set problems.

\section{The design of SAU-Solver}
\subsection{Universal Expression Tree (UET)}
The primary type of textual MWPs can be divided into two groups: arithmetic word problems and equation set problems. For a universal MWPs solver, it is highly demanded to represent various equations of various MWPs in a unified manner so that the solver can generate equations efficiently. Although most of the existing works can handle one-unknown linear word problems well, it is more challenging and harder for current methods to handle the equation set MWPs with multiple unknowns well since they not only do not model the unknowns in the MWPs but also lack of an efficient equations representation mechanism to make their decoder generate required equations efficiently. To handle the above issue, an intuitive way is treating the equation set as a forest of expression trees and all trees are processed iteratively in a certain order. Although this is an effective way to handle equations set problems, it increases the difficulty of equation generation since the model needs to reason out the number of equations before starting equation generation and the prediction error will influence equation generation greatly. Besides, it is also challenging to take full advantage of the context information from the problem and the generated trees. Another way is that we can deploy Seq2Seq-based architecture to handle various equations in infix order like in previous works~\cite{dns,cass}, but there are some limitations, such as generating invalid expression, generating spurious numbers, and generating numbers at wrong positions.

To overcome the above issues and maintain simplicity, we propose a new equation representation called Universal Expression Tree (UET) to make the first attempt to represent the equations of various MWPs uniformly. Specially, we extend the math operator symbol table by introducing a new operator $;$ as the lowest priority operator to integrate one or more expression trees into a universal expression tree, as shown in Fig.~\ref{fig:uet}. With UET, a solver can handle the underlying equations of various textual MWPs easier in a unified manner like the way on arithmetic word problems. Although our UET is simple, it provides an efficient, concise, and uniform way to utilize the context information from the problem and treat the semantic transformation between equations as simple as treating the semantic transformation between subtrees in an equation. 
\subsection{SAU-Solver}
\begin{figure*}[htbp] 
	\centerline{\includegraphics[width=0.99\textwidth]{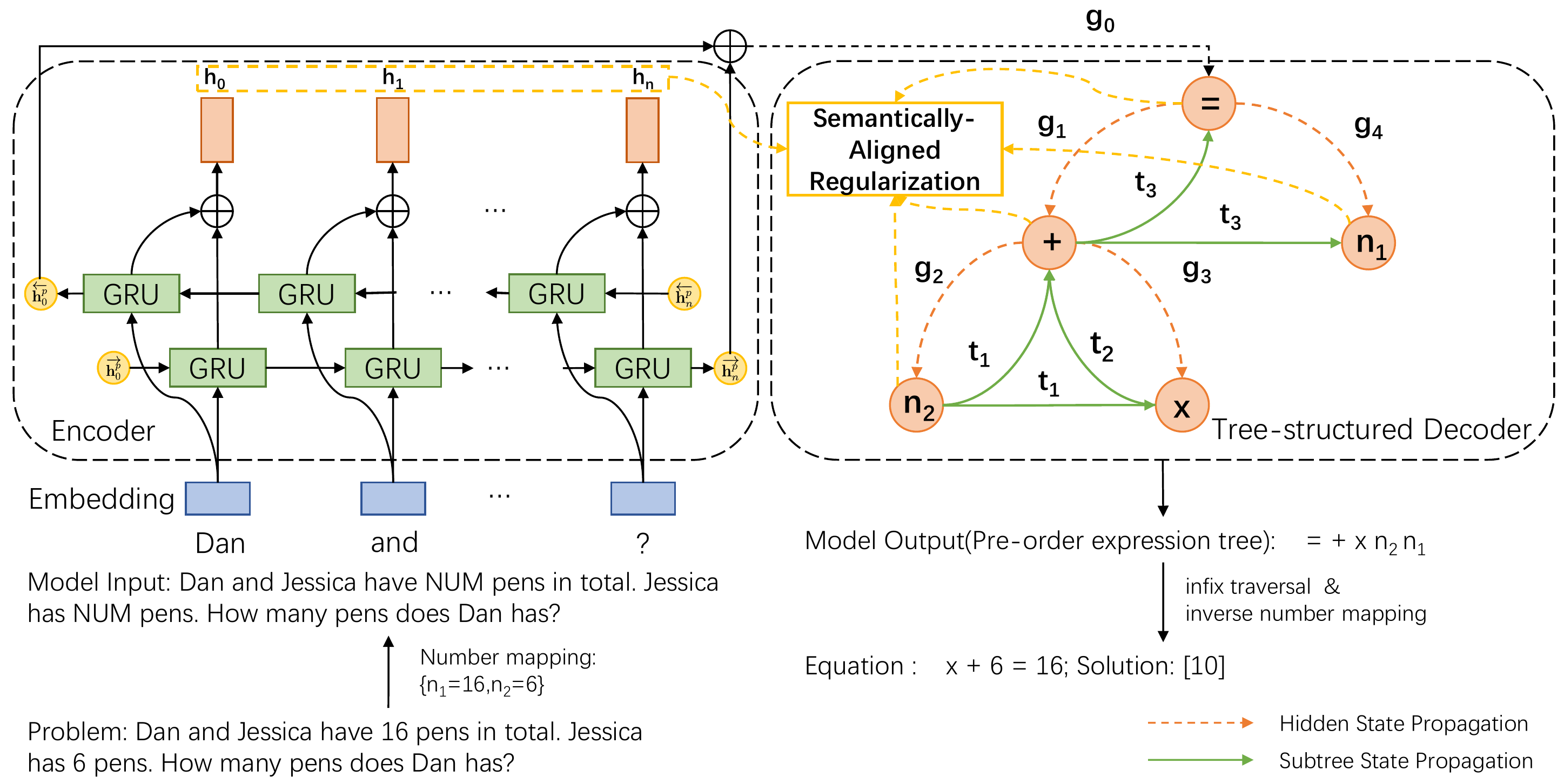}}
	\caption{An overview of our SAU-Solver. When a problem preprocessed by number mapping and replacement is entered, our problem encoder encodes the problem text as context representation. Then our equation decoder generates an expression tree explicitly in pre-order traversal for the problem according to the context representation. Finally, infix traversal and inverse number mapping are applied to generate the corresponding equation.}
	\label{fig:uts}
\end{figure*}
Based on our proposed UET representation, we design a universal tree-structured solver to generate a universal expression tree explicitly according to the problem context and explicitly model the relationships among unknown variables, quantities, math operations, and constants in a tree-structured way, as shown in Fig.~\ref{fig:uts}. Our solver consists of a Bi-GRU-based problem encoder and an explicit tree-structured equation decoder. When a problem is entered, our model first encodes each word of the problem to generate the problem's contextual representation $\mathbf{g}_0$ by our problem encoder. Then, the $\mathbf{g}_0$ will be used as the initial hidden state by our tree-structured equation decoder to guide the equation generation in prefix order with two intertwined processes: top-down tree-structured decoding and bottom-up subtree semantic transformation. With the help of top-down tree-structured decoding and bottom-up subtree semantic transformation, SAU-Solver can generate the next symbol by taking full advantage of generated symbols in a semantically-aligned manner like human solving MWPs. Finally, we apply infix traversal and inverse number mapping to generate the corresponding human-readable equation that can be computed by SymPy\footnote{\url{https://www.sympy.org/}}, which is a python library for symbolic mathematics.

\subsubsection{Problem Encoder}
Bidirectional Gated Recurrent Unit (BiGRU)~\cite{gru} is an efficient method to encode sequential information. Formally, given an input
math word problem sentence $P$ = $\left \{ x_t \right \}^n_{t=1}$, we first embed each word into a vector $\mathbf{x}_t$. Then these embeddings are fed into a two-layer BiGRU from beginning to end and from end to beginning to model the problem sequence:
\begin{equation}\label{1}
\begin{aligned}
&\overrightarrow{\mathbf{h}_{t}^{p}} = GRU(\overrightarrow{\mathbf{h}_{t-1}^{p}}, \mathbf{x}_t) \\
&\overleftarrow{\mathbf{h}_{t}^{p}} = GRU(\overleftarrow{\mathbf{h}_{t+1}^{p}}, \mathbf{x}_t) \\
&\mathbf{h}_{t}^{p} = \overrightarrow{\mathbf{h}_{t}^{p}} + \overleftarrow{\mathbf{h}_{t}^{p}}
\end{aligned}	
\end{equation}
where $GRU(\cdot,\cdot)$ represents the function of a two-layer GRU. $\mathbf{h}_{t}^{p}$ is the sum of the hidden states
$\overrightarrow{\mathbf{h}_{t}^{p}}$  and $\overleftarrow{\mathbf{h}_{t}^{p}}$, which are from both forward and backward GRUs. These representation vectors are then fed into our tree-structured equation decoder for ensemble expression tree generation. Besides, we also 
construct the hidden state $\mathbf{g}_{0}$ as the initial hidden state of our equation decoder:
\begin{equation}\label{2}
\mathbf{g}_{0}^{p} = \overrightarrow{\mathbf{h}_{n}^{p}} + \overleftarrow{\mathbf{h}_{0}^{p}}
\end{equation}
where $\overrightarrow{\mathbf{h}_{n}^{p}}$ and $\overleftarrow{\mathbf{h}_{0}^{p}}$ are the hidden states of forward sequence and backward sequence respectively.

\subsubsection{Equation Decoder}
For decoding, inspired by previous works~\cite{seq2tree,stackdecoder}, we build a semantically-aligned tree decoder to decide which symbol to generate by taking full advantage of the semantic meanings of the generated symbols with two intertwined processes: top-down tree-structured decoding and bottom-up subtree semantic transformation. Our decoder takes tree-based information $\mathbf{g}_{parent}$ (left node) or ($\mathbf{g}_{parent}$, $\mathbf{t}_l$) (right node) as the input and maintains two auxiliary stacks $G$ and $T$ to enforce semantically-aligned decoding procedure. The stack $G$ maintains the hidden states generated from the parent node while the stack T helps the model decide which symbol to generate by maintaining subtree semantic information of generated symbols. Benefiting from UET, our decoder can automatically end the decoding procedure without any special token. If the predicted token $y_t$ is an operator, then we generate two children hidden states $\mathbf{g}_l$ and $\mathbf{g}_r$ according to the current node embedding $\mathbf{n}$ of $y_t$, and push them into the stack $G$ to maintain the state transition among nodes and be used to predict token and its node embedding. Besides, we also push the token embedding $\mathbf{e}(y_t|P)$ of $y_t$ into the stack $T$ so that we can maintain subtree semantic information of generated symbols after right child node generation. If the predicted token $y_t$ is not an operator, we check the size of the stack $T$ to judge whether the current node is a right node. If the current node is a right node, we transform the embedding of parent node $op$, left sibling node $l$ and current node $\mathbf{e}(y_t|P)$ to a subtree semantic representation $\mathbf{t}$, which represents the semantic meanings of generated symbols for current subtree and is used to help the right node generation of the upper subtree. In this way, our equation decoder can decode out an equation as a human writes out an equation according to the problem description.

\noindent\textbf{Token Embedding.} For a problem $P$, its target vocabulary $V^{tar}$ consists of 4 parts: math operators $V_{op}$, unknowns $V_{u}$, constants $V_{con}$ that are those common-sense numerical values occurred in the target expression but not in the problem text (e.g. a chick has 2 legs.), and the numbers $n_p$ occurred in $P$. For each token $y$ in $V^{tar}$, its token embedding $\mathbf{e}(y|P)$ is defined as:
\vspace{-3mm}
\begin{equation}\label{3}
\mathbf{e}(y|P) = \left\{ \begin{array}{ll}
\mathbf{M}_{op}(y) & {\text { if } y \in V_{op}}\\
\mathbf{M}_{u}(y) & {\text { if } y \in V_{u}}\\
\mathbf{M}_{con}(y) & {\text { if } y \in V_{con}}\\
\mathbf{h}_{loc}^{p}(y, P) & {\text { if } y \in n_{P}}\\
\end{array} \right.\\
\end{equation}
where $\mathbf{M}_{op}$, $\mathbf{M}_{u}$, and $\mathbf{M}_{con}$ are three trainable word embedding matrices independent of the specific problem. However, for a numeric value in $n_P$, we take the corresponding hidden state $\mathbf{h}_{loc}^{p}$ from encoder as its token embedding, where $loc(y,P)$ is the index position of numeric value y in $P$.

\noindent\textbf{Gating Mechanism and Attention Mechanism.} To better flow important information and ignore useless information, we apply a gating mechanism to generate node state $\mathbf{n}$ which will be used for predicting the output and generating child hidden states $\mathbf{g}_l$ and $\mathbf{g}_r$ for descendant nodes if the output of the current node is a math operator:
\begin{equation}\label{4}
\begin{aligned} 
q &=\sigma\left(\mathbf{W}_{q} I\right) \\ 
Q &=\tanh \left(\mathbf{W}_{Q} I\right) \\ 
O &= q \odot Q
\end{aligned}
\end{equation}
where $O$ can be a left node state $\mathbf{n}_l$, a right node state, a left child hidden state $\mathbf{g}_l$, or a right child hidden state $\mathbf{g}_r$. For $\mathbf{n}_l$, $I$ is $\mathbf{g}_l$ generated by the parent node. For $\mathbf{n}_r$, $I$ is $[\mathbf{g}_r, \mathbf{t}_l]$ which is the concatenation of the hidden state $\mathbf{g}_r$ generated by the parent node and the subtree semantic embedding $\mathbf{t}_l$ of left sibling. For $\mathbf{g}_l$ and $\mathbf{g}_r$, $I$ is $[\mathbf{n}, \mathbf{c}, \mathbf{e}(y_t|P)]$ which is the concatenation of the current node state $\mathbf{n}$, the contextual vector $\mathbf{c}$ aggregating relevant information of the problem as a weighted representation of the input tokens by attention mechanism, and the token embedding $\mathbf{e}(y_t|P)$ of the predicted token $y_t$. 

For better predicting a token $y_t$ by utilizing contextual information, we deploy an attention mechanism to aggregate relevant information from the input vectors. Formally, given current node state $\mathbf{n}$ and the encoder outputs $\left \{ \mathbf{h}^p_t \right \}^n_{t=1}$, we calculate the contextual vector $\mathbf{c}$ as follows:
\begin{equation}\label{5}
\mathbf{c} = \sum_{s} \frac{\exp \left(\mathbf{V}_{a} \tanh \left(\mathbf{W}_{a}\left[\mathbf{n}, \mathbf{h}_{s}^{p}\right]\right)\right)}{\sum_{i} \exp \left(\mathbf{V}_{a} \tanh \left(\mathbf{W}_{a}\left[\mathbf{n}, \mathbf{h}_{i}^{p}\right]\right)\right)} \mathbf{h}_{s}^{p}
\end{equation}
Based on the contextual vector $\mathbf{c}$ and current node state $\mathbf{n}$, we can predict the token $y_t$ as follows:
\begin{equation}\label{6}
y =\arg \max \frac{\exp (\mathbf{s}(y | \mathbf{n}, \mathbf{c}, P))}{\sum_{i} \exp \left(\mathbf{s}\left(y_{i} | \mathbf{n}, \mathbf{c}, P\right)\right)}
\end{equation}
where 
\begin{equation}\label{7}
\mathbf{s}(y | \mathbf{n}, \mathbf{c}, P)=\mathbf{V}_{n} \tanh \left(\mathbf{W}_{s}[\mathbf{n}, \mathbf{c}, \mathbf{e}(y | P)]\right)
\end{equation}

\noindent\textbf{Subtree Semantic Transformation.} Although our decoder decodes a universal expression tree in the prefix, to help our model to generate the next symbol in a semantically-aligned manner by taking full advantage of the semantic meanings of the generated expression tree, we design a recursive neural network to transform the semantic representations of the current node and its two child subtrees $\mathbf{t}_l$ and $\mathbf{t}_r$ into a high-level embedding $\mathbf{t}$ in a bottom-up manner. Formally, let $t$ be a subtree, and $y$ denotes the predicted token of the root node of the subtree. If $y$ is a math operator, which means that the current subtree $t$ must have two child subtrees $\mathbf{t}_l$ and $\mathbf{t}_r$, the high-level embedding $\mathbf{t}$ should fuse the semantic information from the operator token $y$, the left child subtree $\mathbf{t}_l$ and the right child subtree $\mathbf{t}_r$ as follows:
\begin{equation}\label{8}
\begin{aligned}  
&g_{t}=\sigma\left(\mathbf{W}_{g t}\left[\mathbf{t}_{l}, \mathbf{t}_{r}, \mathbf{e}(\hat{y} | P)\right]\right) \\  
&C_{t}=\tanh \left(\mathbf{W}_{c t}\left|\mathbf{t}_{l}, \mathbf{t}_{r}, \mathbf{e}(\hat{y} | P)\right]\right) \\
&\mathbf{t} = g_{t} \odot C_{t}  
\end{aligned}
\end{equation}
Otherwise, $\mathbf{t}$ is the embedding $\mathbf{e}(y|P)$ of the predicted token $y$ because $y$ is a numeric value, an unknown variable, or a constant quantity and the recursion stops. 

\subsubsection{Semantically-Aligned Regularization}
When a subtree $t$ is produced by our model, this means that we have a computable unit. The semantics of this computable unit should be consistent with the problem text $P$. To achieve this goal, we propose a subtree-level semantically-aligned regularization to help train a better model with higher performance. For each subtree embedding $\mathbf{t}$ and encoder outputs $\left \{\mathbf{h}_1^P,\mathbf{h}_1^P,\cdots, \mathbf{h}_n^P \right \}$, we first apply an attention function to compute a semantically-aligned vector $\mathbf{a}$ as Equation(\ref{5}), then we use a two-layer feed-forward neural network with tanh activation to transform $\mathbf{t}$ and $\mathbf{a}$ into same semantic space respectively. The procedure can be formulated as:
\begin{equation}\label{9}
\begin{aligned}
\mathbf{e}_{sa} &=\mathbf{W}_{e2} \tanh \left( \mathbf{W}_{e1} \mathbf{a} \right) \\
\mathbf{d}_{sa} &=\mathbf{W}_{d2} \tanh \left( \mathbf{W}_{d1} \mathbf{t} \right) \\ 
\end{aligned}
\end{equation}
where $\mathbf{W}_{e1}$, $\mathbf{W}_{e2}$, $\mathbf{W}_{d1}$, and $\mathbf{W}_{d2}$ are trainable parameter matrices. 
With the vectors $\mathbf{e}_{sa}$ and $\mathbf{d}_{sa}$
Let $m$ be the number of subtrees in a universal expression tree, we can regularize our model by minimizing the following loss:
\begin{equation}\label{10}
\mathcal{L}_{sa}(T|P) = \frac{1}{m}\sum_{i=1}^{m}\left\|\mathbf{d}_{sa} - \mathbf{e}_{sa} \right\|_2
\end{equation}

\subsubsection{Training Objective}
Given the training dataset $\mathbf{D}$=\{$(P^i,T^1)$, $(P^2,T^2)$, $\cdots$,$(P^N,T^N)$ \}, where $T^i$ is the universal expression tree of problem $P^i$, we minimize the following loss function:
\begin{equation}\label{12}
\mathcal{L}(T|P) = \sum_{(P,T)\in \mathbf{D}}[-\operatorname{log} p(T|P) + \lambda *\mathcal{L}_{sa}(T|P)]
\end{equation}
where
\begin{equation}\label{13}
p(T|P) = \prod_{t=1}^{m}\operatorname{prob}(y_t|\mathbf{g}_t,\mathbf{c}_t, P)
\end{equation}
where $m$ denotes the size of T, and $\mathbf{g}_t$ and $\mathbf{c}_t$ are the hidden state vector and its contextual vector at the t-th node. We set $\lambda$ as 0.01 empirically. 

\subsection{Discussion}
The methods most relevant to our method are GTS~\cite{seq2tree} and StackDecoder~\cite{stackdecoder}. However, compared with them, our method is different from them as follows. First, our method applies a universal expression tree to represent the diverse equations underlying different MWPs uniformly, which match real-word MWPs better than GTS and StackDecoder which either can only handle single-var linear MWPs without considering unknowns or can handle equations set problem iteratively. Second, we introduce subtree-level semantically-aligned regularization for better enforcing the semantic constraints and rationality of generated expression tree during training, leading to higher answer accuracy, as illustrated in Table~\ref{tab:all}.

\section{Hybrid Math Word Problem Dataset}
Most public datasets for automatic MWPs solving either are quite small such as Alg514~\cite{kushman2014learning}, DRAW-1K~\cite{upadhyay2017annotating}, MaWPS~\cite{koncel2016mawps} or exist some incorrect labels such as Dolphin18K~\cite{huang-etal-2016-well}. An exception is the Math23K dataset which contains 23161 problems labeled well with structured equations and answers. However, it only contains one-unknown linear MWPs, which is not sufficient to validate the ability of a math solver about solving multiple types of MWPs. Therefore, we introduce a new high-quality MWPs dataset, called HMWP, in which each sample is extracted from a Chinese K12 math word problem bank, to validate the universality of math word problem solvers and push the research boundary of MWPs to match real-world scenes better. Our dataset contains three types of MWPs: arithmetic word problems, equations set problems, and non-linear equation problems. There are 5491 MWPs, including 2955 one-unknown-variable linear MWPs, 1636 two-unknown-variable linear MWPs, and 900 one-unknown-variable non-linear MWPs. It should be noticed that our dataset is sufficient for validating the universality of math word problem solvers since these problems can cover most cases about MWPs. We labeled our data with structured equations and answers as Math23K~\cite{dns}. The data statistics of our dataset and several publicly available datasets are shown in Table ~\ref{tab:ds}. From the statistics, we can see that the \#AVG EL (average equation length), \#Avg PN (average number of quantities occurred in problems and their corresponding equations), and \#Avg Ops (average numbers of operators in equations) are the largest among the serval publicly available datasets. ~\cite{seq2tree} showed the higher these values, the more difficult it is. Therefore, our dataset is more challenging for MWPs solvers. 
\begin{table*}[htbp]
\Huge
\centering
\resizebox{0.99\textwidth}{!}{
\begin{tabular}{|c|c|c|c|c|c|c|c|c|c|}
\hline
Dataset & \# Problems & \# Templates & \# Sentences & \# Words &\# Avg EL & \# Avg SNI & \# Avg Constants & \# Avg Ops & Problem types \\ \hline
Alg514 & 514 & 28 & 1.62k & 19.3k  &9.67 &3.54 & 0.44 &5.69 & algebra, linear \\ \hline
Dolphin1878 & 1,878 & 1,183 & 3.30k & 41.4k  &8.18 & 2.58 & 0.63 &4.97  & linear + nonlinear\\ \hline
DRAW-1K & 1,000 & 230 & 6.23k & 81.5k &9.985 &3.386 &0.747 &5.852 &algebra, linear\\ \hline
MaWPS & 2373 &-	&2373 &73.3k	&4.55	&2.31	&0.26	&1.78 &algebra, linear\\ \hline
Math23K & 23,161 & 2,187 & 70.1k & 822k &5.55 &3.0 & 0.28 &2.28 & algebra, linear \\ \hline
Dolphin18k & 18,460 & 5,871 & 49.9k & 604k &9.19 &3.15 & 1.09 &4.96 & linear + nonlinear\\ \hline
HMWP & 5470 & 2779 & 9.56k & 342k &\textbf{10.73} &3.42 &\textbf{1.35} & \textbf{5.96 }& linear + nonlinear\\ \hline 
\end{tabular}}
\caption{Statistics of our dataset and several publicly available datasets. Avg EL, Avg SNI, Avg Constants, and Avg Ops represent average equation length, average number of quantities occurred in problems and their corresponding equations, average numbers of constants only occurred in equations, and average numbers of operators in equations, respectively. The higher these values, the more difficult it is. This has been shown in ~\cite{seq2tree}.}
\label{tab:ds}
\end{table*}

\section{Experiments}
\subsection{Experimental Setup and Training Details}
\textbf{Datasets, Baselines, and Evaluation metric.} We conduct experiments on four datasets, such as HMWP, Alg514~\cite{kushman2014learning}, Math23K~\cite{dns} and Dolphin18K-Manual~\cite{huang-etal-2016-well}. The data statistics of four datasets are shown in Table \ref{tab:ds}. The main state-of-the-art learning-based methods to be compared are as follows: \textbf{Seq2Seq-attn w/ SNI}~\cite{dns} is a universal solver based on the seq2seq model with significant number identification(SNI). \textbf{GTS}~\cite{seq2tree} is a goal-driven tree-structured MWP solver only for one-unknown-variable non-linear MWPs. \textbf{StackDecoder}~\cite{stackdecoder} is a semantically-aligned MWPs solver. \textbf{SAU-Solver w/o SSAR} and \textbf{SAU-Solver} are two universal tree-structured solvers proposed in this paper without and with subtree semantically-aligned regularization. Following our baselines, we use \emph{answer accuracy} as the evaluation metric: if the calculated value of the predicted expression tree equals to the true answer, it is thought of correct since the predicted expression is equivalent to the target expression.
\begin{table}[htbp] 
\Huge
\centering
\resizebox{0.99\linewidth}{!}{
\begin{tabular}{|c|c|c|c|c|}
\hline
Model & HMWP & ALG514 & Math23K & \tabincell{c}{Dolphin18K\\ Manual}\\ \hline
Seq2Seq-attn w/ SNI & 23.2\% & 16.1\% & 58.1\% & 5.9\% \\ \hline
GTS & - &- & 73.9\% & - \\ \hline
StackDecoder & 27.4\%  & 28.86\% & 66.0\% & 9.8\%\\ \hline
SAU-Solver w/o SSAR (ours) & 44.40\% &55.44\% & 74.53\%& 11.02\%\\ \hline
SAU-Solver (ours) & \textbf{44.83\%} & \textbf{57.39\%}& \textbf{74.84\%}& \textbf{11.41\%}\\ \hline 
\end{tabular}}
\caption{Model comparison on answer accuracy via 5-fold cross-validation. ``-'' means either the code is not released or the model is not suitable on those datasets.}
\label{tab:all}
\end{table}

\noindent\textbf{Implementation Details.} We use PyTorch\footnote{http://pytorch.org} to implement our model on Linux with NVIDIA RTX 2080Ti. All the words with less than five occurrences are converted into a special token UNK. We set the dimensionality of word embedding and the size of all hidden states for other layers as 128 and 512, respectively. But for HMWP and Dolphin18K-Manual, we set the size of all hidden states for other layers as 384 since the memory consumption exceeds the capacity of NVIDIA RTX 2080Ti. Our model is trained by ADAM optimizor~\cite{adam} with $\beta_1$ = 0.9, $\beta_2$ =0.999, and $\epsilon$  = $10^{-8}$. The mini-batch size is set to 32. The initial learning rate is set to $10^{-3}$ and then decreases to half every 20 epochs. To prevent overfitting, we set the dropout probability as 0.5 and weight decay as $1e^{-5}$. Finally, we set beam size as 5 in beam search to generate expression trees.
\begin{table}[htbp] 
\Huge
\centering
\resizebox{0.99\linewidth}{!}{
\begin{tabular}{|c|c|c|c|c|}
\hline
 & \tabincell{c}{linear \\ (One-VAR)} & \tabincell{c}{linear \\ (Two-VAR)}  & \tabincell{c}{non-linear \\ (One-VAR)}  & All \\ \hline
\# Num &1944 & 1614 & 1912  & 5470\\ \hline 
\# Avg EL &10.50 &12.10 &9.83  &  10.73\\ \hline
\# Avg SNI &3.59 &3.59 &3.12  & 3.42\\ \hline 
\# Avg Constants &1.21 &1.41 &1.45  &1.35 \\ \hline
\# Avg Ops &5.70 &7.10 &5.26  & 5.96\\ \hline
\tabincell{c}{Correct number \\(Retrieval-Jaccard)}  & 222 &348 & 618  & 1188 \\ \hline
\tabincell{c}{Accuracy \\ ( Retrieval-Jaccard  )}  & 11.42\%   & 21.56\%  & 32.32\%  & 21.72\% \\ \hline
\tabincell{c}{Correct number \\ (Seq2Seq-attn w/SNI)}  & 244 & 312 & 711  & 1267 \\ \hline
\tabincell{c}{Accuracy \\ (Seq2Seq-attn w/SNI)}  & 12.55\%  & 19.33\%  &  37.19\%  & 23.2\%\\ \hline
\tabincell{c}{Correct number \\ (SAU-Solver (ours))}  & \textbf{593} &\textbf{673} & \textbf{1186}  & \textbf{2452}\\ \hline
\tabincell{c}{Accuracy \\ (SAU-Solver (ours))}  & \textbf{30.50\%} & \textbf{41.70\%} & \textbf{62.03\%}  & \textbf{44.83\%}\\
\hline
\end{tabular}}
\caption{The data statistics and performance on different subset of HMWP.}
\label{tab:subset}
\end{table}

\subsection{Results and Analyses}
\begin{table*}[htbp] 
\Huge
\centering
\resizebox{0.99\textwidth}{!}{
\begin{tabular}{|l|l|l|}
\hline
\multicolumn{3}{|l|}{ \tabincell{l}{\textbf{Case 1: } \begin{CJK}{UTF8}{gbsn}鸡兔同笼 ， 上数 有 \texttt{NUM}($n_0$ [20]) 个头 ， 下数 有 \texttt{NUM}($n_1$ [50]) 条腿， 可知 鸡 数量 为 多少 ？\end{CJK} ( An unknown number of rabbits and chickens \\ were locked in a cage, counting from the top, there  were \texttt{NUM}($n_0$ [20]) heads, counting from the bottom, there were \texttt{NUM}($n_1$ [50]) feet. How many chickens were \\ locked in this cage? )}} \\ \hline
\textbf{Seq2Seq:} (x-$n_1$)/$n_0$=(x-$n_1$)/$n_2$; (\textbf{error}) &
\textbf{SAU-Solver w/o SSAR:} $n_0$+$x$+4.0*$x$=$n_1$; (\textbf{correct}) &
\textbf{SAU-Solver:} 2.0*$x$+4.0*($n_0$-$x$)=$n_1$; (\textbf{correct})\\ \hline
\hline
\multicolumn{3}{|l|}{ \tabincell{l}{\textbf{Case 2: }
\begin{CJK}{UTF8}{gbsn}\texttt{NUM}($n_0$ [1]) 艘 轮船 航行 于 A 、 B \texttt{NUM}($n_1$ [2]) 个 码头 之间 ， 顺水 需 \texttt{NUM}($n_2$ [5]) 小时 ，逆水 需 \texttt{NUM}($n_3$ [7]) 小时 ， 已知 水流 速度 为 每\end{CJK} \\ 
\begin{CJK}{UTF8}{gbsn} 小时 \texttt{NUM} ($n_4$ [5]) 千 米 ， 则 A 、 B 之间 距离 为 多少 千 米 ？ \end{CJK} ( \texttt{NUM}($n_0$ [1]) boat sailing between \texttt{NUM}($n_1$ [2]) docks, it takes \texttt{NUM}($n_2$ [5]) hours to sail \\ from A to B downstream, while \texttt{NUM}($n_3$ [7]) hours sailing upstream.  Knowing the velocity of the water flow is 5 km/h, what is the distance between A and B? )}}\\ \hline
\textbf{Seq2Seq:} $x$/($n_2$+$n_1$)+$n_1$=$x$-/$n_2$; (\textbf{error})& \textbf{SAU-Solver w/o SSAR:} $x$/$n_2$-$n_4$=$x$/$n_3$+$n_4$; (\textbf{correct})& \textbf{SAU-Solver:} $x$/$n_2$-$n_4$=$x$/$n_3$+$n_4$; (\textbf{correct})\\ \hline
\hline
\multicolumn{3}{|l|}{ \tabincell{l}{\textbf{Case 3: }
\begin{CJK}{UTF8}{gbsn}整理 \texttt{NUM}($n_0$ [1])  批 图书 ， 如果 由 \texttt{NUM}($n_1$ [1]) 个 人 单独 做 , 要 花 \texttt{NUM}($n_2$ [60]) 小时 ．现在 由 一部分 人 用 \texttt{NUM}($N_3$ [1]) 小时 整理 , 随后 \end{CJK}\\ 
\begin{CJK}{UTF8}{gbsn} 增加 \texttt{NUM}($n_4$[15]) 人 和 他们 一起 又 做 了 \texttt{NUM}($n_5$ [2]) 小时 , 恰好 完成 整理 工作 ．假设 每个 人 的 工作效率 相同 ，那么 先 安排 整理 的 人员 有 \end{CJK}\\
\begin{CJK}{UTF8}{gbsn}多少 人 ？ \end{CJK}  ( Given \texttt{NUM}($n_0$ [1]) stack of books, \texttt{NUM}($n_1$ [1]) student can sort them in \texttt{NUM}($n_2$ [60]) hours. In the first \texttt{NUM}($N_3$ [1]) hours, there were several \\
students sorting books, later, \texttt{NUM}($n_4$[15]) more students joined them, and they finished the job in another \texttt{NUM}($n_5$ [2]) hours together. If each student is as \\ efficient as the others, how many students were working at the beginning?}} \\ \hline
\textbf{Seq2Seq:} $n_1$*($x$/$n_2$)+$n_5$*($x$+$n_4$)/$n_2$=1.0; (\textbf{error}) &
\textbf{SAU-Solver w/o SSAR:} $x$/$n_2$+$n_5$*($x$+$n_4$)/$n_2$=1.0; (\textbf{correct}) &
\textbf{SAU-Solver:} $x$/$n_2$+$n_5$*($x$+$n_4$)/$n_2$=1.0; (\textbf{correct})\\ \hline
\hline
\multicolumn{3}{|l|}{ \tabincell{l}{\textbf{Case 4: } 
\begin{CJK}{UTF8}{gbsn}某 农场 老板 准备 建造 \texttt{NUM}($n_0$ [1]) 个 矩形 羊圈 ，他 打算 让 矩形 羊圈 的 \texttt{NUM}($n_1$ [1]) 面 完全 靠 墙 ，墙 可 利用 的 长度 为 \texttt{NUM}($n_2$ [25]) \end{CJK}\\
\begin{CJK}{UTF8}{gbsn} m ，  另外 \texttt{NUM}($n_3$ [1]) 面 用 长度 为 \texttt{NUM}($n_4$ [50]) m 的 篱笆 围成 ( 篱笆 正好 要 全部 用 完 ， 且 不 考虑 接头 的 部分 ) ，若要 使 矩形 羊圈 的 面积\end{CJK}\\
\begin{CJK}{UTF8}{gbsn} 为 \texttt{NUM}($n_5$ [300]) m \^{} \texttt{NUM}($n_6$ [2]) ， 求 垂直于 墙 的 边 长．\end{CJK} ( A farm owner plans to build a rectangle sheepfold, with \texttt{NUM}($n_1$ [1]) side against the wall. \\ The wall is 25 meters long, and he used \texttt{NUM}($n_3$ [1]) \texttt{NUM}($n_4$ [50])-meter-long fence to build the rest of the sheepfold (the fence should be exactly used up, \\ neglecting the joining part). If the area of the sheepfold is \texttt{NUM}($n_5$ [300]) m \^{} \texttt{NUM}($n_6$ [2]), find the length of the side vertical to the wall.}}\\ \hline
\textbf{Seq2Seq:} $x$*($n_3$-2.0*$x$)=$n_4$; (\textbf{error}) & 
\textbf{SAU-Solver w/o SSAR:} ($n_2$-2.0*$x$)*($n_4$-2.0*$x$)= $n_5$; (\textbf{error}) & 
\textbf{SAU-Solver:} x*($n_4$-2.0*x)= $n_5$; (\textbf{correct})\\ \hline
\hline
\end{tabular}}
\caption{Typical cases. Note that the results are represented as infix traversal of expression trees which is more readable than prefix traversal.}
\label{tab:cs}
\end{table*}
\noindent\textbf{Answer Accuracy.} We conduct 5-fold cross-validation to evaluate the performances of baselines and our models on all four datasets. The results are shown in Table \ref{tab:all}. Several observations can be made from the results in Table \ref{tab:all} as follows: 

First, our SAU-Solver has achieved significantly better than the baselines on four datasets. It proves that our model is feasible for solving multiple types of MWPs. It also proves that our model is more general and more effective than other state-of-the-art models on the real-word scenario that need to solve multiple types of MWPs with a unified solver. 

Second, with our subtree-level semantically-aligned regularization on training procedure, our SAU-Solver has gained additional absolute 0.43\% accuracy on HMWP, absolute 1.95\% accuracy on ALG514, absolute 0.31\% accuracy on Math23k, and absolute 0.39\% accuracy on Dolphin18k-Manual. This shows that subtree-level semantically-aligned regularization is helpful for improving subtree semantic embedding, resulting in improving expression tree generation, especially for the generation of the right child node. Although StackDecoder can be a universal math word problem solver via simple operator extension, the performances on HMWP, ALG514, and Dolphin18k-Manual are very poor, since it generates expression trees independently and only considers the semantic-aligned transformation in an expression tree. Different from it, our SAU-Solver generates multiple expression trees as a universal expression tree and conducts subtree-level semantic-aligned transformation for subsequent tree node generation in our universal expression tree. In this way, we can deliver the semantic information of the previous expression tree to help the generation of the current expression tree. Therefore we can achieve better performance than StackDecoder. 

Overall, our model is more general and effective than other state-of-the-art models on multiple MWPs and outperforms the compared state-of-the-art models by a large margin on answer accuracy.

\noindent\textbf{Performance on different types of MWPs.} We drill down to analyse the performance of \textbf{Retrieval-Jaccard}, \textbf{Seq2seq-attn w/SNI}, and \textbf{SAU-Solver} on different types of MWPs in HMWP. The data statistics and performance results are shown in Table \ref{tab:subset}. We can observe that our model outperforms the other two models by a large margin on all subsets. Intuitively, the longer the expression length is, the more complex the mathematical relationship of the problem is, and the more difficult it is. And the average expression length of our dataset is much longer than Math23K according to the data statistics of Table~\ref{tab:subset} and Table~\ref{tab:ds}. Therefore, we can observe that the accuracy of our model on linear (One-VAR) is lower than Math23K in Table \ref{tab:all}.
\begin{table}[htbp]
\Huge
\centering
\resizebox{0.99\linewidth}{!}{
\begin{tabular}{|c|c|c|c|c|c|c|}
\hline
\multirow{2}{*}{\tabincell{c}{Expression \\ Tree Sizes}}  & \multicolumn{3}{c|}{Math23K} &\multicolumn{3}{c|}{HMWP}  \\ \cline{2-7}
 & Correct & Error & Acc(\%) & Correct & Error & Acc(\%) \\ \hline
3- &729 &168 &81.27\%  &0 &0 &0\% \\ \hline
5 &1872 &435 &81.14\% &3 &1 &75.00\% \\ \hline
7 &620 &291 &68.06\% &32 &25 &56.14\% \\ \hline
9 &147 &143 &50.69\% &159 &69 &69.74\% \\ \hline
11 &66 &74 &47.14\% &102 &111 &47.89\% \\ \hline
13+ &20 &66 &23.26\% &197 &395 &33.28\% \\ \hline
\end{tabular}}
\caption{Accuracy of different expression tree size.}
\label{tab:error}
\end{table}

\subsection{Error Analysis}
In Table~\ref{tab:error}, we show the results of how the accuracy changes as the expression tree size becomes larger. We can observe that as the expression tree size becomes larger, our model's performance becomes lower. This shows that although our model can handle various equations in a unified manner, it still has limitations at predicting long equations since longer equations often match with more complex MWPs which are more difficult to solve. Thus, our model still has room for improvement in reasoning, inference, and semantic understanding. Besides, compared with performances on Math23K which has only a few examples with complex templates, our model achieves significant improvement on the subset of HMWP with expression tree size 13+. This shows that constructing datasets with abundant complex examples can improve the model's ability to handle complex problems. 

\subsection{Case Study}
Further, we conduct a case analysis and provide four cases in Table~\ref{tab:cs}, which shows the effectiveness of our approach. Our analyses are summarized as follows. From \textbf{Case 1}, Seq2Seq generates a spurious number $n_2$ not in problem text while both SAU-Solver w/o SSAR and SAU-Solver predict correctly owning to the problem-specific target vocabulary. Besides, although both SAU-Solver w/o SSAR and SAU-Solver can generate correct an equation, the equation generated by our SAU-Solver is more semantically-aligned with a human than the equation generated by SAU-Solver. From \textbf{Case 2}, we can see that Seq2Seq generates an invalid expression containing consecutive operators while our models can guarantee the validity of expressions since they generate expression trees directly. From \textbf{Case 3}, we find it interesting that tree-based models can avoid generating redundant operations, such as ``$n_1$*''. From \textbf{Case 4}, we can see that SAU-Solver can prevent generating the similar subtree as its left sibling when the parent node is ``*''. 

\section{Conclusion}
We propose an SAU-Solver, which is able to solve multiple types of MWPs, to generate the universal express tree explicitly in a semantically-aligned manner. Besides, we also propose a subtree-level semantically-aligned regularization to improve subtree semantic representation. Finally, we introduce a new MWPs datasets, called HMWP, to validate our solver's universality and push the research boundary of MWPs to math real-world applications better. Experimental results show the superiority of our approach.

\section*{Acknowledgements}
We thank all anonymous reviewers for their constructive comments. This work was supported in part by National Key RD Program of China under Grant No. 2018AAA0100300, National Natural Science Foundation of China (NSFC) under Grant No.U19A2073 and No.61976233, Guangdong Province Basic and Applied Basic Research (Regional Joint Fund-Key) Grant No.2019B1515120039, Nature Science Foundation of Shenzhen Under Grant No. 2019191361, Zhijiang Lab’s Open Fund (No. 2020AA3AB14), Sichuan Science and Technology Program (No. 2019YJ0190). 

\bibliography{treedecoder}
\bibliographystyle{acl_natbib}
\end{document}